\documentclass{article}

\usepackage{arxiv}

\usepackage[utf8]{inputenc} % allow utf-8 input
\usepackage[T1]{fontenc}    % use 8-bit T1 fonts
\usepackage{hyperref}       % hyperlinks
\usepackage{url}            % simple URL typesetting
\usepackage{booktabs}       % professional-quality tables
\usepackage{amsfonts}       % blackboard math symbols
\usepackage{nicefrac}       % compact symbols for 1/2, etc.
\usepackage{microtype}      % microtypography
\usepackage{cleveref}       % smart cross-referencing
\usepackage{lipsum}         % Can be removed after putting your text content
\usepackage{graphicx}
\usepackage{natbib}
\usepackage{doi}
\usepackage{multirow}
\usepackage{color}
\newcommand{\link}[1]{{\color{blue}\href{#1}{#1}}}

\title{StyleT2F: Generating Human Faces from Textual Description Using StyleGAN2}

\author{{Mohamed Shawky Sabae}\\
	Department of Computer Engineering\\
	Cairo University\\
	Cairo, Egypt \\
	\texttt{mohamed.sabae99@eng-st.cu.edu.eg} \\
	\And
	{Mohamed Ahmed Dardir}\\
	Department of Computer Engineering\\
	Cairo University\\
	Cairo, Egypt \\
	\texttt{mohamed.dardir98@eng-st.cu.edu.eg} \\
	\And
	{Remonda Talaat Eskarous}\\
	Department of Computer Engineering\\
	Cairo University\\
	Cairo, Egypt \\
	\texttt{remonda.bastawres99@eng-st.cu.edu.eg} \\
	\And
	{Mohamed Ramzy Ebbed}\\
	Department of Computer Engineering\\
	Cairo University\\
	Cairo, Egypt \\
	\texttt{mohamed.ibrahim98@eng-st.cu.edu.eg} \\
}

\renewcommand{\shorttitle}{StyleT2F: Generating Human Faces from Textual Description Using StyleGAN2}

\hypersetup{
pdftitle={StyleT2F: Generating Human Faces from Textual Description Using StyleGAN2},
pdfsubject={text2face, gans},
pdfauthor={Mohamed S. Sabae, Mohamed A. Dardir, Remonda T. Eskarous, Mohamed R. Ebbed},
pdfkeywords={Generative adversarial networks (GANs), Text-to-face generation, Latent manipulation},
}

\begin{document}
\maketitle

\begin{abstract}
	AI-driven image generation has improved significantly in recent years. Generative adversarial networks (GANs), like \emph{StyleGAN} \cite{karras2019stylebased}, are able to generate high-quality realistic data and have artistic control over the output, as well. In this work, we present \emph{StyleT2F}, a method of controlling the output of \emph{StyleGAN2} \cite{karras2020analyzing} using text, in order to be able to generate a detailed human face from textual description. We utilize \emph{StyleGAN}'s latent space to manipulate different facial features and conditionally sample the required latent code, which embeds the facial features mentioned in the input text. Our method proves to capture the required features correctly and shows consistency between the input text and the output images. Moreover, our method guarantees disentanglement on manipulating a wide range of facial features that sufficiently describes a human face. Our code and results are available at: \link{https://github.com/DarkGeekMS/Retratista}
\end{abstract}

% keywords can be removed
\keywords{Generative adversarial networks (GANs), Text-to-face generation, Latent manipulation}

\begin{figure}[ht]
 \centering
  \includegraphics[width=\textwidth]{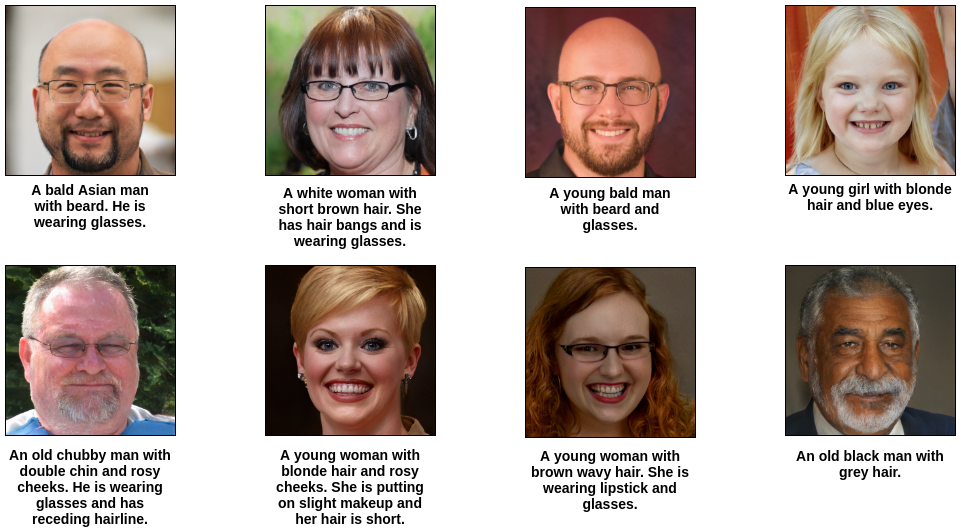}
  \caption{Visual samples showing the consistent mapping between input textual keywords and generated facial features.}
  \label{fig:visuals}
\end{figure}

\section{Introduction}
Recent advances in deep generative models have enabled the generation of high-dimensional visual data. Generative adversarial networks (GANs) have been one of the most successful generative models, as they not only generate accurate high-quality data but also structure their latent representation accurately, so that different features are independent of each others. Traditional GANs, such as \emph{DCGAN} \cite{radford2016unsupervised}, yield low-dimensional results and no control over output features. However, recent techniques enable precise manipulation of different data features, which is crucial for many applications. \emph{StyleGAN} \cite{karras2019stylebased} and \emph{BigGAN} \cite{brock2019large} are two popular examples of these models for high-resolution image generation. In this work, we study the latent space of \emph{StyleGAN2} \cite{karras2020analyzing} and propose a method for sampling latent embeddings of human faces, conditioned by text. Our work extends other research work \cite{abdal2019image2stylegan} \cite{shen2020interfacegan} that targets the manipulation of \emph{StyleGAN} latent space. Our method targets consistent mapping between the features of the input text and the generated face image, while targeting a wide range of facial features.

\section{Related Work}
\emph{Generative modelling} is one the most active areas of research in the Machine Learning community. Numerous works focus on visual data generation, especially images. Generating synthetic data can be useful for training data augmentation, scientific studies and more. Conditional generative models can be even more useful, as data samples are conditionally generated based on specific input, such as text. Generative adversarial networks (GANs) \cite{goodfellow2014generative} are wide used for visual data generation and offer robust performance due to adversarial loss.

\subsection{Text-to-Image Generation}
Text-to-Image Generation aims to translate an input text to a visual image, whether it's a face, a natural scenery or generally any scene. To our knowledge, the earliest proposed work for text-to-image generation is \cite{reed2016generative}, where simply text embeddings are concatenated to the noise vector of a GAN image generator. However, the network yields poor results and fails to capture the mapping between input text and output image features. Consequently, more work was done to address these issues. \emph{StackGAN} \cite{zhang2017stackgan} and \emph{StackGAN++} \cite{zhang2018stackgan} proposed a hierarchical image generation using a pair of generators and discriminators. \emph{AttnGAN} \cite{xu2017attngan} included an attention mechanism that successfully matches input text with the corresponding image features. Recently, \emph{DALL-E} \cite{ramesh2021zeroshot} introduces a zero-shot text-to-image generation utilizing \emph{CLIP} \cite{radford2021learning} and variational autoencoders (VAEs). \emph{FuseDream} \cite{liu2021fusedream} utilizes GAN architectures with well-structured latent space, such as \emph{BigGAN} \cite{brock2019large}, for general image generation from text. \emph{StyleCLIP} \cite{patashnik2021styleclip} uses \emph{StyleGAN} \cite{karras2019stylebased} \cite{karras2020analyzing} along with \emph{CLIP} \cite{radford2021learning} to manipulate a human face using text.

\subsection{The Latent Space of GANs}
A latent space is an embedding space, where high-dimensional data are encoded in a structured way, such that similar features are grouped together. Conventionally, a GAN generates images from simple random noise sampled from a normal distributions. However, more recent well-trained GAN architectures offer a structured latent space that disentangles different features from each other. Many recent architectures are proposed to offer a disentangled latent space. \emph{StyleGAN} \cite{karras2019stylebased} \cite{karras2020analyzing} is one of the most widely-used architectures. Consequent work \cite{abdal2019image2stylegan} \cite{shen2020interfacegan} \cite{2021} attempted to study \emph{StyleGAN} latent space and extract disentangled directions that can manipulate different image features independently.

Our work aims to extend this even further to be able to completely generate a high resolution detailed human face from an input textual description. We try to include as many facial features as possible to enable accurate and comprehensive description of a human face.

\section{Face Generation from Text}
\begin{figure}[ht]
\centering
  \includegraphics[width=\textwidth]{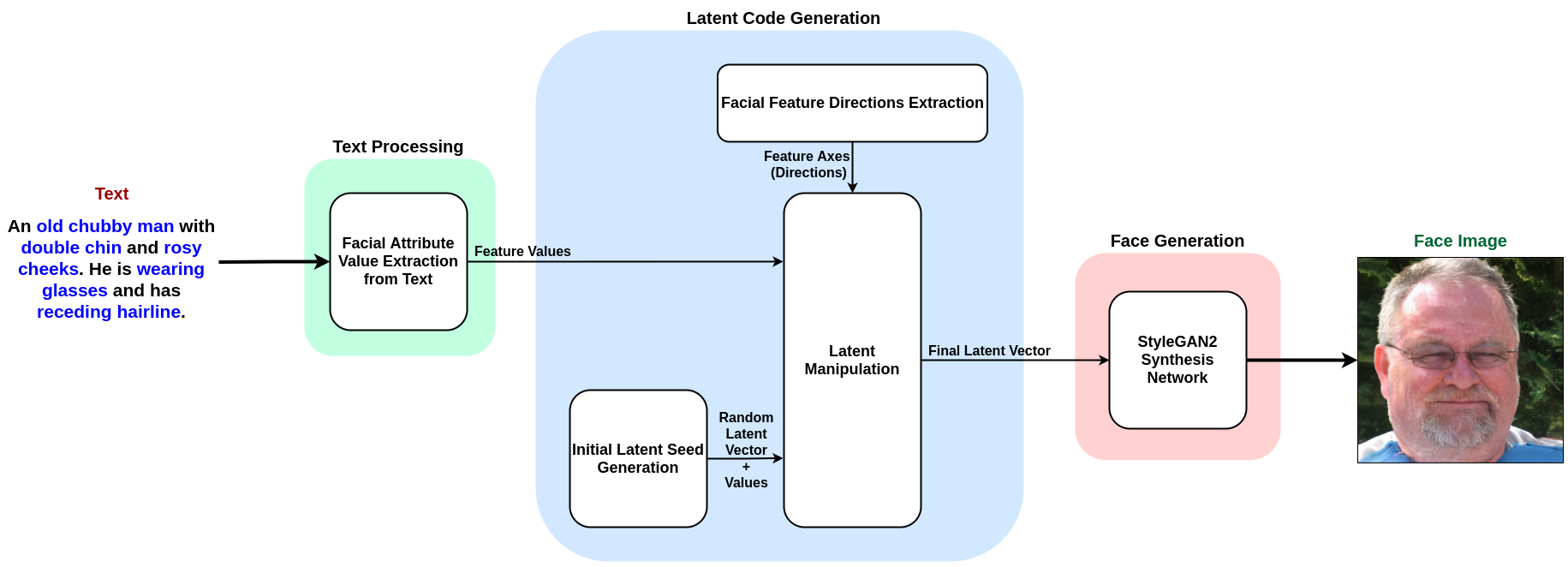}
  \caption{Face generation from text system overview. The system consists of 3 stages. First, the input text is processed to extract the facial attribute values. Then, these feature values are used to manipulation StyleGAN2 latent space, in order to sample the latent code that represents such features. Finally, the extracted latent code is passed to StyleGAN2 synthesis network to generate the final face image.}
  \label{fig:overview}
\end{figure}

\subsection{Method Overview}
We propose a complete pipeline for human face generation from textual description. As shown in \ref{fig:overview}, the input description is first passed to a \emph{text processing} module that extracts the required facial features and gives a value (score) to each of them. The target feature values are then passed to \emph{face code generation} module that extracts the correct latent embedding corresponding to such values. Finally, we use \emph{StyleGAN2} \cite{karras2020analyzing} synthesis network to generate the complete human face image from the extracted latent vector. Our work is mainly an analysis of \emph{StyleGAN2} latent space and integration of text processing module for face generation from text.

\subsection{Text Processing}
The first step is to extract the required facial feature values, which are later encoded in the latent vector, from the input text. To formulate our problem, we have to answer two main questions; [1] What features should we target to correctly describe a human face? [2] How should we numerically encode the scale of such features? To address the first question, we have to come up with a set of facial features that comprehensively model the human face. We decide on $32$ features, including hair color, eye color, facial hair and more (refer to \ref{appendix:features} for more details). For the second question, we empirically induce a certain range of values for each feature that suits \emph{StyleGAN2} latent space navigation. For example, \emph{"A man with heavy beard"} should have a higher score for facial hair feature than that of \emph{"A man with beard"}. Also, refer to \ref{appendix:features} for more details. Thus, the goal of text processing module is to encode text into $32$ values (logits) corresponding to the considered facial features, each of them has a specific numeric value that represents its level. Consequently, our problem can be formulated as \emph{multi-label classification}, however we are concerned with the actual logits not just the classification. To solve such a problem, we use \emph{DistilBERT} \cite{sanh2020distilbert}, a transformer-based network, which is relatively compact and performs well. We are left with one last problem, which is the dataset. To our knowledge, there are no current datasets that directly target human face description. To overcome this problem, we pseudo-generate a dataset for the training purpose (more details are shown in \ref{subsec:texttrain}). The resulted network can process the input textual description into the required feature values.

\subsection{Latent Code Generation}
Now, we discuss the methodology of converting feature values into a latent embedding that fit \emph{StyleGAN2} latent space. The general idea is that we start at an initial random latent vector and then navigate in \emph{StyleGAN2} latent space using certain feature directions, in order to reach the target latent vector. This process can be broken down into three sub-modules, discussed below.

\subsubsection{Feature Directions Extraction}
\begin{figure}[ht]
\centering
  \includegraphics[width=0.8\textwidth]{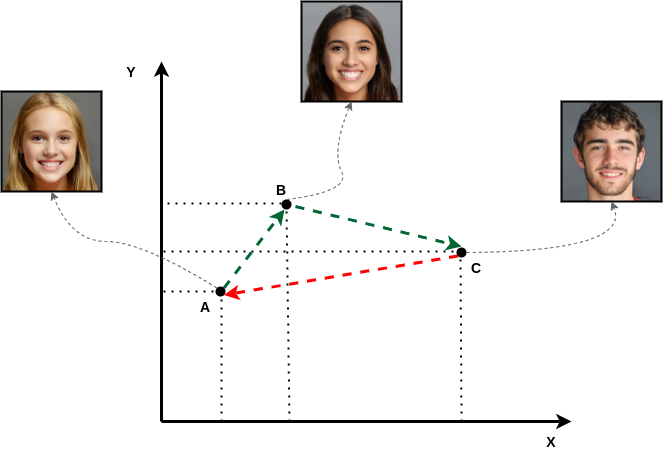}
  \caption{A simple example of a 2D latent space, where x and y are the bases. The axis AB represents the direction of hair color feature, while the axis BC represents the direction of gender feature. We can see that moving along CA should invert the effect of moving along AB then BC.}
  \label{fig:latent}
\end{figure}
We start by extracting the directions in \emph{StyleGAN2} latent space for the required facial features. These directions disentangle a specific feature from others, in order to independently manipulate this feature. Figure \ref{fig:latent} further illustrates the idea of feature directions and latent navigation. We follow a similar methodology to that described in \cite{abdal2019image2stylegan} and \cite{shen2020interfacegan}, however we extend the extracted feature directions to the $32$ considered facial features. The extraction process goes as follows:
\begin{itemize}
    \item We sample a number of face images, generated by \emph{StyleGAN2}, with their corresponding latent vectors $w+$.
    \item We then categorize the sampled images based on each facial features. This categorization can be based on a discrete value (e.g. with or without beard) or a continuous value (e.g. hair length). For this step, we use various facial attributes recognition methods (mentioned in \ref{appendix:classifiers}). The obtained data forms a latent vector to facial feature values correspondence, from which we can extract latent directions for each feature (directions that can manipulate specific features in the latent space independently). 
    \item Finally, we use \emph{logistic regression} (in case of discrete labels) and \emph{linear regression} (in case of continuous labels) to learn the feature directions.
\end{itemize}
Thus, the latent direction of each of the $32$ facial features is extracted and can be used to manipulate this specific feature.

\subsubsection{Initial Seed Generation}
As mentioned above, we start from an initial random latent vector, then use the feature directions to reach the latent vector that satisfies the target feature values. To do so, we have to sample a random latent vector $w+$ and get the feature values corresponding to it. We sample an initial $z$ vector from a standard normal distribution, then pass it through \emph{StyleGAN2} mapping network to get the initial $w+$ vector. Furthermore, to get the corresponding feature values, we simple get the component of the latent vector along each feature direction using \emph{dot product}:
\begin{equation}
    v_f = L_{rand} \cdot \hat{D}_f
\end{equation}
Where $v_f$ is the value of feature $f$, $L_{rand}$ is the initial random latent vector and $\hat{D}_f$ is the unit vector of the feature direction. Thus, we have an initial latent vector, along with its feature values.

\subsubsection{Latent Manipulation}
Lastly, we generate the target latent vector using \emph{latent manipulation}. Once the initial latent latent vector is generated, we use feature directions to reach a latent vector that satisfies the target feature values. We do so using sequential navigation along each feature direction. In other word, we navigate along the feature directions one by one to reach the target feature values from the current values. We do not consider the directions of the features that are not mentioned in the input description. The following equation summarizes our navigation method in a vectorized form:
\begin{equation}
    L_{target} = L_{rand} + (V_{target} - V_{rand}) * \textbf{D}
\end{equation}
Where $L_{target}$ is the target latent vector, $L_{rand}$ is the initial random latent vector and $V_{target}$ and $V_{rand}$ are $32D$ vectors corresponding to the target and initial feature values, respectively. $(V_{target} - V_{rand})$ represents the difference between target and initial values, thus the amount of required navigation. Keep in mind that the unmentioned features are not considered (difference set to $0$). $\textbf{D}$ is the feature directions matrix.

Note that to get more accurate results, we re-project the latent vector, after each navigation, on all feature directions to get the updated feature values. This compensates for any shift that occurs during sequential navigation due to potential feature directions entanglement. Thus, we have the final latent vector that corresponds to the required feature values.

\subsection{Face Generation}
The final stage is to translate the final latent vector, extracted from \emph{latent manipulation}, to the target human face image. We pass the final latent vector $w+$ to \emph{StyleGAN2} synthesis network that generates the face image. We experiment with \emph{StyleGAN2} latent space, until we reach a good representation of the facial features.

\section{Experimental Setup}
\subsection{Datasets}
For this work, we use two datasets. The first one is the text dataset, which consists of textual face descriptions with their corresponding facial feature labels. Each description has 32 labels corresponding to the facial features, each indicates whether the feature is mentioned or not and its scale (value), if mentioned. To our knowledge, there is no such publicly-available dataset. Consequently, we have to generate a training dataset ourselves. First, we handcraft some descriptions with their corresponding labels manually. We include different sentence lengths to improve generalization. After that, we paraphrase each sentence multiple times, in order to increase the diversity of the dataset. We use \emph{cycle translation} to do paraphrasing, where we translate the sentence from English to another language and then back to English. Thus, we have a diverse dataset that suits our needs. The second dataset is the one used to learn the feature directions. As mentioned above, we sample a number of images and then categorize them based on each facial feature using facial feature classifiers (refer to \ref{appendix:classifiers}).

\subsection{Implementation Details}
We use \emph{DistilBERT} as the network for text processing. While being compact and fast, it still gives accurate results in our case. Moreover, we sample about $3000$ face images from \emph{StyleGAN2} to fit the feature directions. The feature directions are extracted from the extended latent space $w+$, so that we don't need the mapping network during generation. We use the full \emph{StyleGAN2} synthesis network, which generates images at the resolution of $1024X1024$. Consequently, the latent vectors are of $18X512$ dimensions.

\subsection{Text Feature Extraction Training}
\label{subsec:texttrain}
\emph{DistilBERT} is trained on our synthetic dataset of about $40000$ descriptions using \emph{mean-squared error} (MSE) loss. We use \emph{ADAM} optimizer with learning rate of $0.001$ and a batch size of $64$. We train for $30$ epochs using the pretrained weights as initialization, which takes about $12$ hours on an \emph{Nvidia GTX 1080ti}. 

\section{Discussion and results}
\begin{table}[ht]
\caption{Angles (measured in degrees) between different feature directions using a subset of the considered facial features (closer to $90$ degrees is better).}
\centering
\begin{tabular}[t]{| c | c | c | c | c |}
\hline
\textbf{Angles} & Age & Gender & Beard & Gray Hair \\
\hline \hline
Age & 0.0 & 92.4 & 85.8 & 79.6 \\
\hline
Gender & 92.4 & 0.0 & 80.0 & 88.6 \\
\hline
Makeup & 88.0 & 107.7 & 100.5 & 94.5 \\
\hline
Hair Length & 89.7 & 95.9 & 90.6 & 96.6 \\
\hline
\end{tabular}
\label{tab:angles}
\end{table}

\begin{figure}[ht]
\centering
  \includegraphics[width=0.8\textwidth]{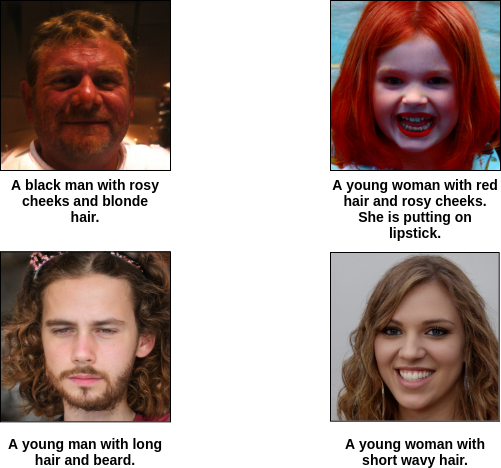}
  \caption{Generation failure cases due to different reasons. The failure can occur due to contradicting input features or uncommon features combination, as in the top images. Also, it can occur due to sequential latent navigation and entangled features, as in the bottom images.}
  \label{fig:failures}
\end{figure}

Our proposed pipeline provides a consistent mapping between textual description and face images. Figure \ref{fig:visuals} shows visual results of our system, where we can see that the input facial features are captured correctly in the output face. Also, table \ref{tab:angles} shows the angles between some feature directions. We can see that most angles are close to $90$ degrees, which shows that the directions are disentangled. However, some feature directions are still naturally entangled, such as gender and makeup. Unfortunately, our system still suffers from multiple failure cases that can arise due to contradicting input facial features, sequential latent manipulation or excessive navigation on certain directions. Figure \ref{fig:failures} shows multiple failure cases for different reasons. Moreover, the whole text-to-face generation pipeline takes about $1$ second on an \emph{Nvidia GTX 1080ti}.

\section{Conclusion}
This work presented a complete pipeline for human face generation from textual description. We utilized the power of \emph{StyleGAN2} and the subsequent work that studied its latent space to build our method. Our method offers a consistent mapping between the input text and the generated images, as well as controlling a set of facial features that sufficiently describes a human face. However, following the same methodology we described, more fine-detailed facial features can be considered to better describe the face. 

\bibliographystyle{unsrtnat}
\bibliography{paper}

\appendix
\section{Considered Facial Features}
\label{appendix:features}

\begin{table}[ht]
\centering
\caption{List of considered facial features (attributes) along with their types.}
\begin{tabular}[t]{| c | c | c |}
\hline
Group & Feature & Type \\
\hline \hline
Eyebrows & Bushy eyebrows & Discrete \\
\hline
\multirow{5}{*}{Hair color} &  Black hair & Discrete \\
& Red hair & Discrete \\
& Blonde hair & Discrete \\
& Brown hair & Discrete \\
& Gray hair & Discrete \\
\hline
\multirow{5}{*}{Hair style} & Curly-straight hair & Continuous \\
& Receding hairline & Discrete \\
& Baldness & Discrete \\
& Hair bangs & Discrete \\
& Hair length & Continuous \\
\hline
Facial hair & Beard & Continuous \\
\hline
\multirow{2}{*}{Race} & Asian & Discrete \\
& Skin color & Continuous \\
\hline
\multirow{10}{*}{General facial attributes} & Face thickness & Continuous \\
& Gender & Discrete \\
& Age & Continuous \\
& Lips size & Continuous \\
& Nose size & Continuous \\
& Ears size & Continuous \\
& Double chin & Discrete \\
& High cheekbones & Discrete \\
& Pointy nose & Discrete \\
& Rosy cheeks & Discrete \\
\hline
\multirow{6}{*}{Eyes} & Black eyes & Discrete \\
& Green eyes & Discrete \\
& Blue eyes & Discrete \\
& Brown eyes & Discrete \\
& Eye size & Continuous \\
& Eye bags & Discrete \\
\hline
\multirow{2}{*}{Makeup} & Makeup saturation & Continuous \\
& Lipstick & Discrete \\
\hline
\multirow{2}{*}{Eyeglasses} & Sight glasses & Discrete \\
& Sun glasses & Discrete \\
\hline
\end{tabular}
\label{tab:features}
\end{table}

Table \ref{tab:features} shows the considered facial features along with their type, whether discrete or continuous. Note that we split the colors (e.g. hair color and eye color) into several feature, in order to easily extract the required feature directions and have better control over them. We group similar features together for better representation.

\section{Latent Navigation using Feature Directions}
\begin{figure}
\centering
  \includegraphics[width=\textwidth]{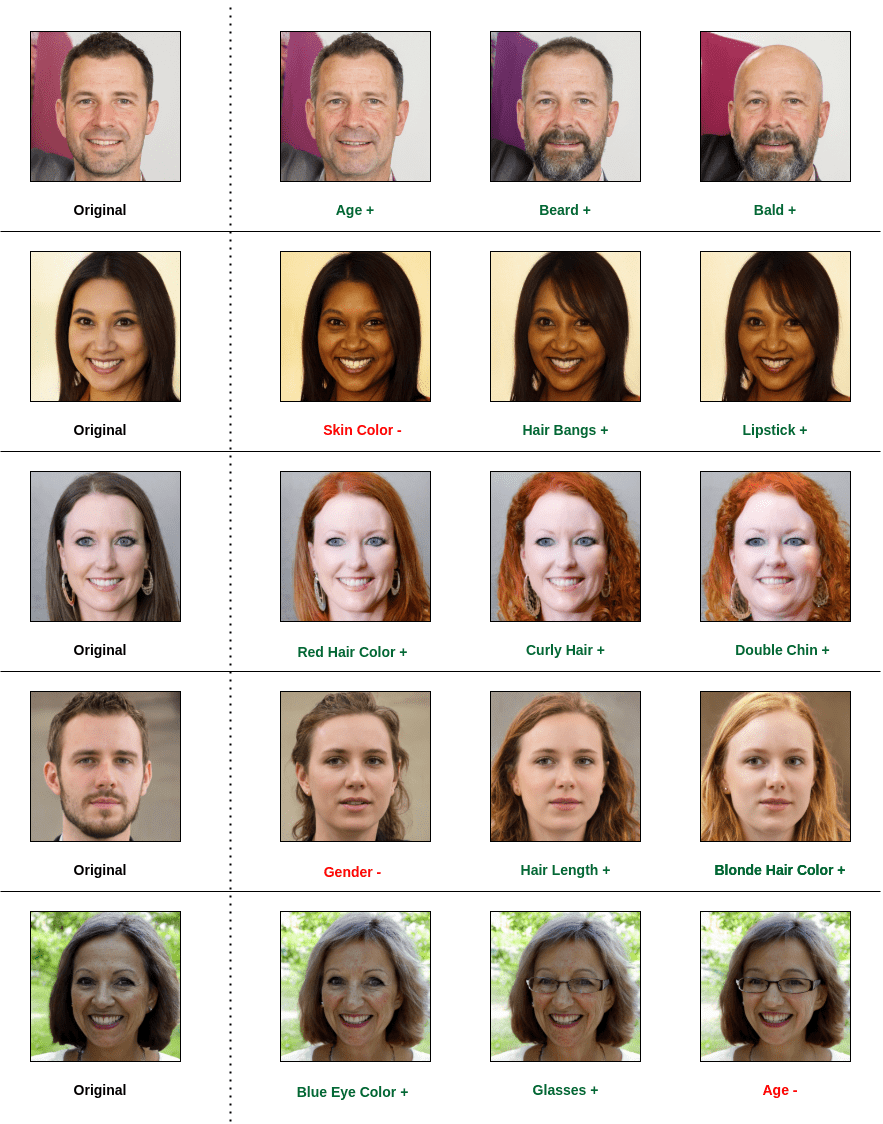}
  \caption{The results of sequential navigation over multiple feature directions. The visual results show the disentanglement between different feature directions, where moving along a certain direction can independently manipulate a single facial feature with little to no changes in the others.}
  \label{fig:navigation}
\end{figure}

We ensure that the extracted feature directions are disentangled as much as possible from each other. We conduct a visual study to assess such disentanglement. Figure \ref{fig:navigation} shows the visual results of sequential navigation over multiple feature directions starting from different seeds. Moving along a certain feature direction can manipulate this specific facial feature with minimum changes in the other features.

\section{Facial Attributes Classifiers}
\label{appendix:classifiers}
To label the synthetic face images based on different facial features, we have to use a range of methods. These methods can be summarized as follows:
\begin{itemize}
    \item \textbf{Manual labelling} is the first idea to come to our minds, where we tried to manually categorize synthetic faces according to certain features. This process is tedious, so we used it just for verification later on.
    \item \textbf{Classical image processing techniques} are, also, used to categorize synthetic faces based on some features. Mainly, we used these techniques to detect colors like eye and hair color. We use \emph{morphological operators} and \emph{classical segmentation} to detect eyes or hair and retrieve their colors.
    \item \textbf{Deep learning techniques} (\emph{neural networks)} are mainly used with features related to facial landmarks, such as eye size. We used \emph{High-Resolution Representations for Labeling Pixels and Regions} \cite{sun2019highresolution} to do \emph{facial landmark detection}, in order to calculate these feature values.
\end{itemize}

\end{document}